# Accelerating Clinical NLP at Scale with a Hybrid Framework with Reduced GPU Demands: A Case Study in Dementia Identification

Jianlin Shi, MD, PhD[1,2], Qiwei Gan, PhD[1,2], Elizabeth Hanchrow, RN, MSN[1, 2], Annie Bowles, MS[1,2], John Stanley, MS[1,2], Adam P. Bress, PharmD, MS[1,2], Jordana B. Cohen, MD, MSCE[3], Patrick R. Alba, MS[1,2]

[1]VA Salt Lake City Health Care System, UT; [2] University of Utah, Salt Lake City, UT, USA; [3] University of Pennsylvania, Philadelphia, PA, USA

**Abstract**

*Clinical natural language processing (NLP) is increasingly in demand in both clinical research and operational practice. However, most of the state-of-the-art solutions are transformers-based and require high computational resources, limiting their accessibility. We propose a hybrid NLP framework that integrates rule-based filtering, a Support Vector Machine (SVM) classifier, and a BERT-based model to improve efficiency while maintaining accuracy. We applied this framework in a dementia identification case study involving 4.9 million veterans with incident hypertension, analyzing 2.1 billion clinical notes. At the patient level, our method achieved a precision of 0.90, a recall of 0.84, and an F1-score of 0.87. Additionally, this NLP approach identified over three times as many dementia cases as structured data methods. All processing was completed in approximately two weeks using a single machine with dual A40 GPUs. This study demonstrates the feasibility of hybrid NLP solutions for large-scale clinical text analysis, making state-of-the-art methods more accessible to healthcare organizations with limited computational resources.*

**Introduction**

The increasing adoption of electronic health records (EHR) in healthcare has led to a rise in clinical data on a variety of scales. A large proportion of these data are in the format of free text, which requires efficient processing techniques to make it computationally usable.[1] While prior research has primarily focused on enhancing the accuracy of natural language processing (NLP) systems, the critical aspect of processing efficiency and cost has not been paid enough attention in the academic domain. The growing demand for large language models requires substantial computational resources[2], making these advanced techniques less accessible to many healthcare organizations.

Various methods, such as quantization,[3] distillation,[4] and pruning,[5] can significantly reduce processing time, but not on a magnitude scale. Although previous studies have investigated hybrid approaches, most have been focused primarily on improving accuracy rather than accelerating processing for large datasets.[6–8] We recognize the additional potential of applying a hybrid approach to enhance execution speed. In this work, we propose a framework that combines a rule-based strategy with a Support Vector Machine (SVM) and a BERT model through a strategically orchestrated process, which mitigates the disadvantages of each method while preserving competitive accuracy. We demonstrate its effectiveness through a case study to identify incident dementia in a cohort of patients with newly diagnosed hypertension for a retrospective clinical study with a total of 4.9 million patients.

Previous studies have demonstrated that relying solely on structured International Classification of Diseases (ICD) codes to identify dementia is insufficient. Historically, NLP has been investigated mostly via rule-based approaches[9], although an increasing number of studies now adopt machine-learning-based methods. Because these studies do not share the exact cohort definition, task specification, and measurements, the reported performance scores are not strictly comparable. In general, machine learning-based studies have demonstrated higher sensitivity and specificity. However, even the best ensemble method has been applied only to cohorts in the thousands, reporting precision and recall of 0.9 and 0.94[10]. In a study with a sample size of 5.4 million patients, a LightGBM model was used to predict future dementia onset, yielding sensitivity between 0.25 and 0.45, and specificity between 0.26 and 0.68.[11] Another study applied logistic regression to detect dementia patients from a 44,674-patient cohort, and reported a sensitivity of 0.772 and specificity of 0.998. In our case study, we employ the proposed hybrid NLP strategy to identify incident dementia among 4.9 million veterans with incident hypertension, drawing on 2.1 billion clinical notes. The resulting cohort will be used to evaluate the long-term effects of new use of antihypertensive medications—angiotensin receptor blockers (ARBs) versus angiotensin-converting enzyme inhibitors (ACEIs)—on dementia risk. This research aims to address critical gaps in understanding whether certain antihypertensive medication classes offer greater dementia-related benefits independent of, or in addition to, their blood pressure–lowering effects. By extending the findings of previous studies, our work will provide actionable insights for high-risk populations to prevent or delay onset of dementia.

**Methods**
*Cohort derivation*
We defined a cohort of patients with incident hypertension (HTN) who received care within the Veterans Health Administration (VHA) network and newly initiated an angiotensin receptor blocker (ARB)– or angiotensin-converting enzyme inhibitor (ACEI)–based regimen (i.e., new users). The index period, determined by data availability, spanned from January 1, 2000, to December 31, 2022. The index date was defined as the date of the first ARB or ACEI prescription fill within this period. To ensure new-user status, we excluded patients with any antihypertensive medication fills in the preceding year. Patients who received both an ARB and an ACEI within a 30-day window were excluded. Hypertension was identified by the presence of at least one outpatient ICD-9/10 code (401.x, I10, I12.0, or I12.9) in the year prior to the index date. In addition, patients were required to have at least one primary care visit at least one year before the index date. Baseline characteristics—including comorbidities, laboratory values, and other antihypertensive medications—were determined using previously described algorithms and all available encounter data from the pre-index period.

All clinical notes from the defined cohort were used. For downstream analyses, these notes aligned with the cohort's index date were processed finally. However, for NLP development and evaluation, notes were sampled from the study period (which matches the index period), including those recorded before the index date.

*Sampling strategies to create the NLP dataset*
To address the challenge of sparsely distributed dementia-related statements in the dataset, a mixed stratified sampling strategy was adopted. This approach aimed to balance annotation efficiency and coverage. The strategies included:
1. **Keyword Sampling:** 96 keywords curated by domain experts and derived from the literature were used to identify dementia-related notes. These keywords were selected to balance between retrieving sufficient content and maintaining relevance.
2. **Note Type Sampling:** A set of note types was curated with domain experts and refined through preliminary analyses.
3. **Specialty Sampling:** Similar to note types, a set of clinical specialties was applied for sampling.
4. **Vector Similarity Sampling:** After a predetermined number of annotation rounds, this strategy was applied by calculating the vector distances between unlabeled snippets and already labeled snippets to identify the notes containing "uncertain" snippets.
5. **Random Sampling:** A portion of the notes was selected entirely at random, without other constraints.

A total of 16 rounds of annotations were completed. The first 14 rounds employed strategies 1 through 4 in a 3:2:2:1 ratio, while the final two rounds used all strategies in a 4:1:1:1:3 ratio. For each strategy, patients were sampled first, then 1–4 notes were sampled for each patient.

*Annotation process*
We developed the annotation guideline with the domain experts based on previous studies and DSM-5. The guideline included the following annotation types: Cognitive or Functional Impairment (CFI), Dementia, Dementia medications, Interference with daily living (Interference), Acute Intoxication, Other Cause of Impairment, and three exclusion conditions: Delirium, Significant Psychiatric Disorder (PsychDisorder), Recent Psychiatric Hospitalization. Then, patient-level labels were derived based on the workflow in Figure 1.
We used eHOST[12] to create the annotations. Two annotators with clinical annotation experience and an adjudicator with both clinical and informatics background were trained over three rounds, annotating 40 notes per round. A snippet-level inter-annotator agreement reached 0.93 in the third round. Formal annotation started in the fourth round. Rounds 4 to 10 were fully double-annotated and adjudicated to ensure consistency between the annotators. Rounds 11 to 16 gradually increased the number of notes per round, reaching a maximum of 140 notes per round. In these rounds, notes were partially double annotated to improve annotation efficiency, with only overlapping notes adjudicated. In total, 2152 notes of 1339 patients were sampled and annotated.

*Training the VABERT model*
The initial NLP development was based on a locally pretrained BERT model (VABERT) using 3 million VA clinical notes.[13] First, the annotated data were split by patient IDs into training (1746 notes from 1071 patients) and testing (406 notes from 268 patients). Next, the annotated data were processed using medspaCy[14] to organize the manually annotated clinical concepts into dataframes that included sentence-level context information. The training data were further split into the training and validation datasets to support model training. Because clinical sentences may mention multiple concepts or clinical conditions simultaneously, a multi-label sequence classifier was used, allowing each sentence to receive more than one label. This classifier was trained on a machine equipped with dual A100 GPUs, using 5e-05 learning rate, 0.1 warmup ratio, 64 training batch size, and 128 evaluation batch size.

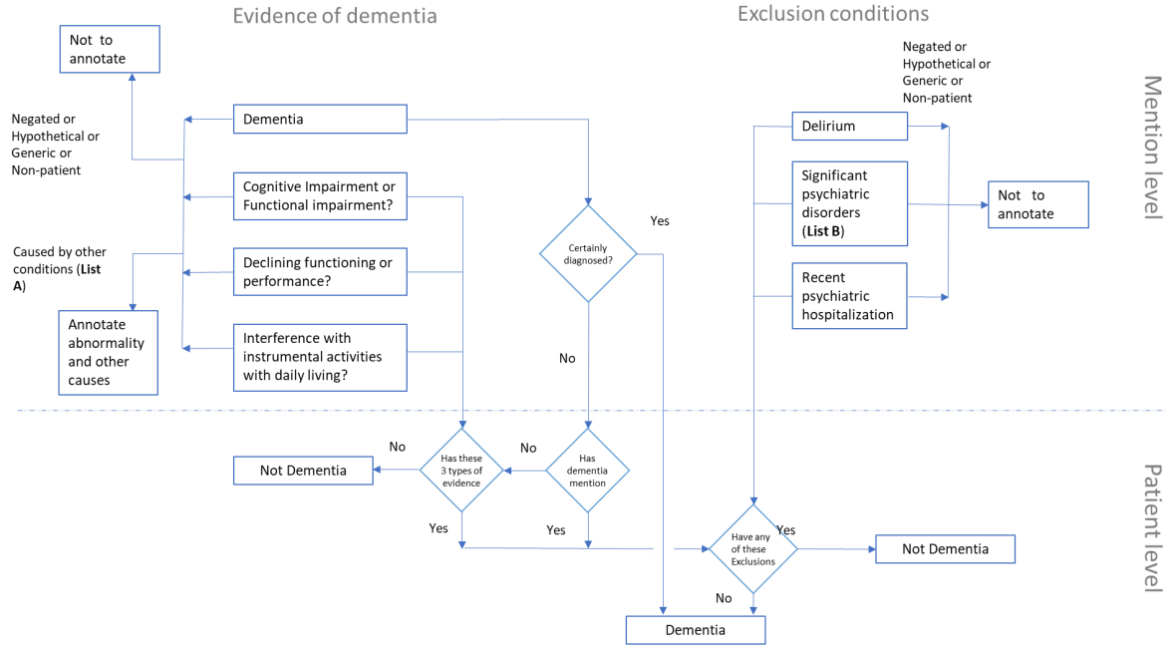

**Figure 1.** Annotation workflow

*Hybrid framework optimization*
The optimization and execution steps are illustrated in Figure 2. First, a list of filtering keywords prioritizing high sensitivity was extracted from the training data and curated manually. (Note: these keywords focused more on recall and differed from those used in the sampling step above.) We verified that using this list of keywords covered all the training annotations. Next, the trained VABERT model was applied to process over 0.8 million randomly sampled notes from the unannotated cohort data. Using the labeled data from the model, 16 specific subwords were identified from note titles to serve as the initial note exclusion filters. Then, sentences identified by the model as containing at least one clinical concept (i.e., labeled as positive or "not-null") were selected for further review; sentences labeled as "null" (containing no concepts), sentences already containing the filtering keywords, or duplicates were excluded. The remaining sentences underwent manual review to identify additional filtering keywords. This process resulted in additional 11 misspellings and a total of 45 filtering keywords.

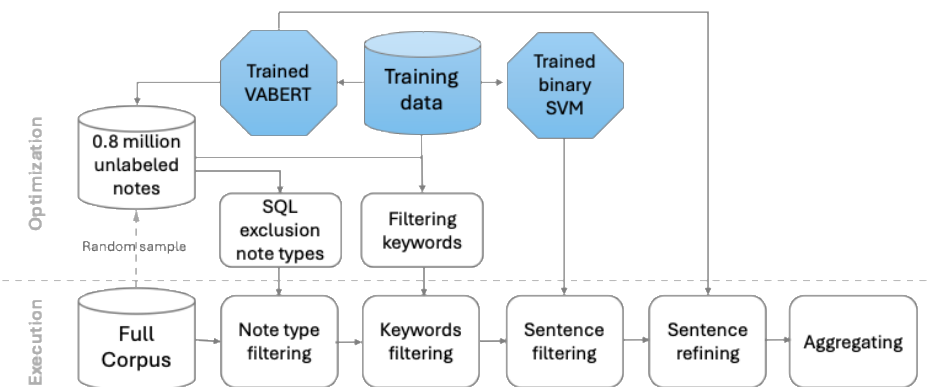

**Figure 2.** The overview of the optimization and execution workflow.

Applying the derived SQL exclusion conditions and keyword filtering to the total cohort data resulted in a 59-million-note corpus. It was estimated that processing this corpus solely with the trained VABERT model on the aforementioned GPU machine would take more than six months. Thus, an additional layer of filtering was added to expedite the process by training a binary sentence SVM classifier using training data to classify whether a sentence contained any annotation. We used the TfIdf vectorizer with linear kernel, and class-balanced weights due to the highly skewed data distribution.

*Hybrid framework execution*
During execution, the SQL exclusion condition with keyword filtering (using SQLServer's full text search function) was first applied to filter the input notes. Next, medspaCy was used to parse each note into sentences, and the trained SVM classifier was employed to further trim the sentences. All these steps were parallelized using PySpark and completed on a non-GPU machine with 64 cores and 2Tb memory. The filtered sentences were shared to the GPU machine via a network drive, where the trained VABERT model was applied to refine the sentence labels. Once all the sentences were processed, a rule-based component was used to aggregate sentence-level predictions to patient-level dementia classifications. Consistent with the annotation guideline logic shown in Figure 1, this component explicitly excluded mentions associated with exclusion conditions (e.g., delirium, acute intoxication, prolonged hospitalization) or attributed to other potential causes (e.g., depression, PTSD, traumatic brain injury, acute substance intoxication, poorly controlled depression). The results were stored into a database table for downstream analyses.

*Evaluation*
The processing time was recorded. Both the trained VABERT and SVM models were evaluated against the test dataset using precision, recall, and F1 measures. The VABERT model was evaluated at both the snippet and patient levels, whereas the SVM model was evaluated only at the snippet level. At patient level, we also compared the performance of the hybrid method with using VABERT model alone. Additionally, we compared the NLP classification with structured codes (including dementia related ICD9 and ICD10 codes, and dementia medication codes, see Appendix). Then, we randomly sampled 80 patients from the execution phase for whom the NLP conclusions did not match the codes (higher probability to detect potential NLP errors than random sampling) and manually reviewed the output.

**Results**
The original corpus consisted of 2.1 billion clinical notes. Applying note type and keywords filtering resulted in 59 million notes and applying the SVM filtering resulted in 129 million sentences. These filtering steps took approximately one week. The sentence refining step using the VABERT model took about eight days, and the aggregating step took four hours.

The binary sentence SVM classifier achieved a precision of 0.76, a recall of 0.93, and an F1 score of 0.84 on the test dataset. Table 1 shows the snippet-level performance of the VABERT model. At the patient level, the hybrid method achieved a precision of 0.90, a recall of 0.84, and an F1 score of 0.87. These scores are even higher than that of the VABERT model alone, which recorded a precision of 0.79, a recall of 0.82, and an F1 score of 0.81 (See Table 2). When comparing the NLP output with structured codes using 80 sampled patients, every patient with a dementia ICD code was identified by the NLP system. Among the patients identified as dementia positive by NLP without structured data support, seven were suspected of having dementia, two were misclassified due to undetected negation, and three were misclassified due to undetected non-patient diagnoses. Compared with the structured codes within the whole corpus, our NLP solution has identified over three times as many dementia patients that met the case study criteria.

**Discussion**
In this study, we proposed a hybrid NLP framework consisting of data-driven note type and keyword filtering, SVM sentence filtering, and a BERT-based model to efficiently process a large corpus of clinical notes (2.1 billion notes). This approach significantly reduced the processing time from an estimated period of more than half a year to just half a month (including around eight days of single GPU machine usage), without compromising accuracy. This framework is not only a cost-efficient solution for large healthcare institutions but also enables small- to medium-sized organizations with limited computational resources to operationalize advanced NLP systems.

Compared with previous studies, this case study is not directly comparable because the study cohort differs. Additionally, our goal is to process real-world data, so we did not limit our note types to generally well‐written ones, such as admission notes and discharge summaries. Moreover, given the heterogeneous nature of note type naming in the VA across 170 medical centers and 1,193 outpatient sites, it is currently not feasible to filter input data using an inclusive list of note types, as there are hundreds of thousands of them.[15] Thus, we used a small set of exclusion conditions to remove certain note types, which were also verified using processed notes. As a result, the NLP task is expected to be more challenging than in prior studies due to the greater variety of note types and structures. Despite this disadvantage, our reported scores rank among the highest when compared to those from previous studies. This suggests that this hybrid framework can effectively bridge the gap between performance and practical feasibility in large-scale healthcare analytics.

**Table 1.** Snippet-level performance of the VABERT model.

| Labels | Precision | Recall | F1 Score | Support |
|---|---|---|---|---|
| CFI | 0.42 | 0.66 | 0.52 | 230 |
| CFI_Worsen | 0.00 | 0.00 | 0.00 | 15 |
| Delirium | 0.43 | 0.46 | 0.44 | 13 |
| Dementia | 0.74 | 0.86 | 0.80 | 64 |
| Dementia_Meds | 0.83 | 0.88 | 0.85 | 43 |
| Interference | 0.07 | 0.06 | 0.07 | 47 |
| Intoxication | 0.13 | 0.19 | 0.16 | 21 |
| OtherCause | 0.30 | 0.36 | 0.33 | 106 |
| PsychDisorder | 0.61 | 0.89 | 0.72 | 28 |
| Hospitalization | 0.00 | 0.00 | 0.00 | 5 |
| **Micro-average** | 0.44 | 0.56 | 0.49 | 572 |

CFI: cognitive functional impairment; CFI_Worsen: worsening CFI; Interference: interference with daily living; PsychDisorder: significant psychiatric disorder; Hospitalization: recent psychiatric hospitalization.

**Table 2**. Patient-level performance in identifying patients with dementia.

| Methods | Precision | Recall | F1 Score | Specificity | Support |
|---|---|---|---|---|---|
| Hybrid | 0.90 | 0.84 | 0.87 | 0.99 | 31 |
| VABERT alone | 0.79 | 0.82 | 0.81 | 0.98 | 31 |

These scores are for dementia-positive patients only. Because the dataset has a skewed distribution—with the majority (237 patients) being dementia-negative—the average performance was even higher, but that measure is not representative of this task.

The success of this framework can be attributed to its multi-layered design. The initial rule-based filtering, which used curated keywords with SQL full-text search, eliminated a large proportion of irrelevant notes. In this case study, 97% of the total notes were removed at this step. However, the actual percentage could vary in different tasks due to the specific keywords list used. The binary SVM classifier further refines the candidate sentences by identifying those that are likely to contain evidence related to dementia, thereby optimizing the use of the resource-intensive VABERT model. This integration allowed us to capitalize on the high sensitivity of traditional approaches while harnessing the superior contextual understanding of modern transformer models.

While this hybrid framework demonstrated robust performance in this case study, error analysis revealed several areas for improvement. A subset of misclassification was due to incorrectly detected context information, which could be caused by insufficient examples in the training set, incorrectly split sentences, or model limitations. Table 1 shows that multiple concepts, including Cognitive or Functional Impairment (CFI) and Interference with daily living (Interference), did not perform well at the snippet level. This was partially due to large language variations between the training and test sets. After a preliminary analysis of the NLP output over the labeled dataset and 0.8 million processed notes, we found that these contributed very few additional true patients while introducing a significant number of false positives. Thus, we decided not to include them in the final aggregation step. "Recent psychiatric hospitalization" and "Worsening CFI" did not perform well and could be evaluated correctly due to their rarity in our annotated dataset.

Future studies might focus on these poorly performed concepts to use more advanced techniques to enrich the data or sampling strategy to improve the performance and make them useful. We may also test different sentence splitting methods. We initially used PyRuSH[16], which yielded more accurate sentence boundaries and resulted in better performance (a precision of 0.89, a recall of 1.00, and an F1 score of 0.94). However, due to execution speed considerations, we used spaCy's sentencizer in the execution phase instead. As a result, some notes contained excessively long sentences, leading to errors. Enhancing the PyRuSH implementation for speed might help future use cases. Looking ahead, validating the framework in other clinical domains and healthcare settings would help assess its generalizability. Finally, although we demonstrate the framework in a retrospective study, integrating this hybrid framework with real-time EHR systems could pave the way for real-time clinical decision support based on timely patient data analysis.

**Conclusion**
We introduced a hybrid NLP framework for large-scale data processing, which combines rule-based note filtering, SVM sentence filtering, and a BERT-based model. Using a case study to identify dementia patients from a corpus of 2.1 billion notes, we demonstrated that the framework reduced processing time from over six months to around two weeks while maintaining competitive accuracy. Future work will focus on refining methods for underrepresented concepts, improving sentence segmentation methods, and validating the framework across diverse clinical settings.

**Acknowledgment**
This work was supported by the NIH-funded project "Informing optimal first-line antihypertensive therapy: A rigorous comparative effectiveness analysis of ARBs vs. ACEIs on long-term risk of dementia, cancer, heart disease, and quality of life" (1R01AG074989-01), the Department of Veterans Affairs funded project "Put VA data to work for veterans" (VA ORD 24-D4v-02). The analysis was conducted within VA informatics and Computing Infrastructure (VINCI) and an AWS VA Enterprise Cloud enclave (Prospect). We would like to thank Dr. Siamack Ayandeh for the creation of the enclave and support of this project and VHA Office of the Research and Development for funding the Cloud Credits.

**Appendix**

The structured data approach for identifying the dementia cohort was based on the following criteria:

ICD codes:

- ICD-9 codes: 290.0, 290.1x, 290.2x, 290.3, 90.4x, 290.8, 290.9, 291.1, 291.2, 292.82, 294.0, 294.10, 294.11, 294.20, 331.0, 331.82, 331.11, 331.19, 333.0, 797

- ICD-10 codes: F06.1, F06.8, F07.0, G13.2, G13.8, G23.0, G23.1, G23.2, G23.8, G23.9, G30.8, G30.9, G31.01, G31.09, G31.1, G31.2, G31.83, G31.85, G31.89, G31.9, G94, R41.81, R54

Medication Criteria:

The cohort also included individuals who have dispensed at least one dementia-related medication, including memantine, galantamine, rivastigmine, or donepezil.